\pdfoutput=1

\documentclass[11pt]{article}

\usepackage[preprint]{acl}
\onecolumn

\usepackage{times}
\usepackage{latexsym}

\usepackage[T1]{fontenc}

\usepackage{inputenc}

\usepackage{microtype}

\usepackage{inconsolata}

\usepackage{graphicx}
\usepackage{amsmath}
\usepackage{enumitem}
\usepackage{booktabs}

\usepackage{authblk}

\title{NormCode: A Semi-Formal Language for Auditable AI Planning}

\author[1,2]{Xin Guan\thanks{Corresponding author: xin.guan@psylensai.com}}
\author[3,1]{Yunshan Li\thanks{2400101083@mails.szu.edu.cn}}
\author[4,1]{Zekun Wu}
\author[4,1]{Ruibo Zhang\thanks{ruibo.zhang.23@ucl.ac.uk}}

\affil[1]{Psylens.AI}
\affil[2]{Center for Long-Term AI}
\affil[3]{Shenzhen University}
\affil[4]{University College London}

\begin{document}

\maketitle

\begin{abstract}
As AI systems move into high-stakes domains---legal reasoning, medical diagnosis, financial decision-making---regulators and practitioners demand auditability: the ability to trace exactly what each step in a multi-step workflow saw and did. Yet current LLM-based workflows are fundamentally opaque. \textit{Context pollution}---the accumulation of information across reasoning steps---causes models to hallucinate and lose track of constraints, while implicit data flow makes it impossible to reconstruct what any given step actually received. We present \textbf{NormCode}, a semi-formal language that makes AI workflows \textit{auditable by construction}. Each inference operates in enforced data isolation with only explicitly passed inputs, eliminating cross-step contamination and ensuring that every intermediate state is inspectable. A strict separation between \textit{semantic operations} (LLM reasoning) and \textit{syntactic operations} (deterministic data flow) enables auditors to distinguish probabilistic inference from mechanical restructuring. The multi-format ecosystem (\texttt{.ncds}, \texttt{.ncd}, \texttt{.ncn}, \texttt{.ncdn}) allows developers, domain experts, and auditors to inspect the same plan in formats suited to their needs. A four-phase compilation pipeline transforms natural language intent into executable JSON repositories, while a visual Canvas App provides real-time graph visualization and breakpoint debugging. We validate the approach through 100\% accuracy on base-X addition and self-hosted execution of NormCode's own compiler, demonstrating that structured intermediate representations can bridge human intuition and machine rigor while maintaining full transparency.
\end{abstract}

\section{Introduction}

Large language models have enabled AI systems that reason, plan, and act across complex multi-step workflows. Frameworks like LangChain, AutoGPT, and AutoGen demonstrate how agents can decompose problems and coordinate actions through sequences of LLM calls. Yet as these systems move into high-stakes domains---legal analysis, medical reasoning, financial decision-making---a fundamental question emerges: \textbf{can we audit what these systems actually did?}

\subsection{The Problem: Opaque Workflows and Unauditable Reasoning}

For AI systems operating in regulated or high-consequence domains, auditability is not optional. The EU AI Act mandates decision traceability for high-impact systems. Professional standards in medicine and law require that practitioners justify their reasoning. Financial regulations demand that algorithmic decisions be explainable~\cite{sai2025}. Yet current LLM-based workflows are fundamentally opaque: when step 7 of a 10-step chain produces an unexpected result, developers cannot easily determine what step 7 actually saw.

This opacity stems from \textbf{context pollution}---the accumulation of information across reasoning steps that causes models to hallucinate, confuse intermediate outputs with inputs, and lose track of original task constraints. Empirical studies show that LLMs ``forget'' earlier facts in long conversations, leading to contradictory decisions on lengthy tasks~\cite{chang2025, saga2025}. The compounding error effect is severe: small mistakes in early steps propagate through sequential reasoning, making complex multi-step solutions exponentially more error-prone than simple ones~\cite{wand2025}.

Even state-of-the-art models like GPT-4 struggle to maintain global consistency on complicated plans, often abandoning earlier constraints as context grows. Simply expanding context windows does not solve the problem---research shows that very long prompts can confuse models or cause them to focus on irrelevant details, actually \textit{reducing} correctness despite larger windows~\cite{breunig2025}.

Current approaches offer no path to auditability. Direct prompting bundles everything into one context window, creating cognitive overload with no separation of concerns. Chain-of-Thought (CoT) prompting extends interaction step-by-step but does not isolate context between steps---hallucinated thoughts in early reasoning leak forward into later prompts. Agent frameworks like LangChain and AutoGPT provide orchestration but leave data flow implicit~\cite{ibm2025, wang2024}. As practitioners note, debugging such agents often happens ``in the dark,'' requiring reverse-engineering of hidden state to find root causes~\cite{patel2025}. If developers cannot reconstruct what happened, auditors certainly cannot.

\subsection{The Solution: Auditability Through Enforced Data Isolation}

NormCode makes AI workflows \textbf{auditable by construction}. The key mechanism is explicit data isolation: rather than letting context bleed implicitly between steps, NormCode defines \textit{plans of inferences}---structured decompositions where each step is a self-contained logical unit with access only to explicitly passed inputs. If an early step processes a raw document and later steps receive only a summarized excerpt, no subsequent inference can accidentally peek at the original---it simply is not in that step's input by design.

This architectural choice produces three forms of auditability. \textbf{Input auditability}: for any step, you can answer ``what exactly did this step see?'' by inspecting its explicitly declared inputs. \textbf{Process auditability}: the separation between semantic operations (LLM reasoning) and syntactic operations (deterministic data flow) lets you trace exactly which steps involved probabilistic inference and which were mechanical restructuring. \textbf{Output auditability}: every intermediate result is stored and inspectable, creating a complete audit trail from inputs to conclusions.

This approach aligns with emerging strategies in advanced agent design. Experts advocate isolating context into separate threads or sub-agents, each handling narrow subtasks with only relevant data, rather than one monolithic agent juggling a combined context~\cite{ruan2024}. Recent work on execution isolation (e.g., IsolateGPT) proposes sandboxing LLM-based applications to prevent unauthorized data access~\cite{wu2025}. NormCode builds this isolation into the language itself---not as a guideline but as an enforced rule of the programming model, guaranteeing that every execution produces a verifiable audit trail.

\subsection{Contributions}

This paper presents NormCode, a semi-formal language and execution framework that makes AI planning auditable by construction. Our contributions are as follows.

First, we introduce \textbf{an auditable intermediate representation} for AI planning based on inference decomposition and explicit data flow. The representation uses tensor-structured references with named axes to organize data flow precisely, ensuring that every step's inputs and outputs are fully traceable.

Second, we establish \textbf{semantic/syntactic separation} as a clean architectural distinction between LLM-driven operations (expensive, non-deterministic) and data-restructuring operations (free, deterministic). This separation enables auditors to distinguish probabilistic reasoning from mechanical data routing, supporting precise cost and reliability attribution.

Third, we provide \textbf{a multi-format ecosystem} (\texttt{.ncds} for draft authoring, \texttt{.ncd} for formal execution, \texttt{.ncn} for natural language verification, \texttt{.ncdn} for hybrid editing) that enables different stakeholders---developers, domain experts, auditors---to inspect the same underlying logic in formats suited to their needs.

Fourth, we describe \textbf{a four-phase compilation pipeline} (Derivation, Formalization, Post-Formalization, Activation) that transforms natural language intent into executable JSON repositories through progressive refinement, with manual review opportunities at each phase.

Fifth, we present \textbf{a working implementation} including an orchestrator with dependency-driven scheduling, SQLite-backed checkpointing, and loop management, as well as a visual Canvas App debugger built on React Flow and FastAPI providing real-time graph visualization, breakpoint debugging, and multi-agent configuration---making the audit trail not just available but visually explorable.

Sixth, we validate the approach through \textbf{two demonstrations}: (a) 100\% accuracy on base-X addition tasks across arbitrary digit lengths, and (b) self-hosted execution of NormCode's own compiler pipeline, demonstrating both correctness and expressive completeness.

The rest of this paper is organized as follows: Section 2 reviews related work in AI planning, agent frameworks, and intermediate representations. Sections 3-4 describe the NormCode language and reference system. Section 5 articulates the design philosophy. Sections 6-7 detail the execution model and compiler ecosystem. Section 8 presents case study evaluations. Section 9 discusses limitations and future work, and Section 10 concludes.

\section{Background and Related Work}

NormCode builds on three research traditions: classical AI planning, modern LLM-based agents, and intermediate representations for structured reasoning. This section positions NormCode within this landscape.

\subsection{Classical Planning and Structured Task Decomposition}

NormCode inherits its hierarchical task decomposition from classical AI planning. The seminal STRIPS system introduced representing actions with preconditions and effects for goal-directed problem solving. The Planning Domain Definition Language (PDDL) emerged as a standardized formalism for encoding planning domains---providing structured symbolic blueprints (predicates, actions, goals) that automated planners use to generate valid action sequences.

Hierarchical Task Network (HTN) planning is particularly relevant to NormCode's design~\cite{erol1994}. HTN represents tasks in a hierarchy where high-level goals are recursively decomposed into lower-level primitive actions. This decomposition mirrors NormCode's approach of breaking complex inferences into hierarchies of sub-inferences, each with explicit dependencies. However, while HTN planning operates in fully symbolic domains with deterministic actions, NormCode extends this paradigm to handle \textit{probabilistic reasoning} (via LLM calls) within a structured framework.

\subsection{Modern LLM-Based Agent Frameworks}

Recent agent frameworks demonstrate how LLMs can plan and act through iterative reasoning. \textbf{ReAct}~\cite{yao2022} interleaves reasoning traces with action outputs, allowing models to ``think aloud'' while interfacing with tools. \textbf{Reflexion}~\cite{shinn2023} extends this with self-evaluation: after each trial, the agent generates linguistic feedback stored as episodic memory, enabling iterative self-correction without parameter updates.

Frameworks like \textbf{LangChain} and \textbf{AutoGPT} provide orchestration for multi-step LLM workflows~\cite{ibm2025}. \textbf{LangGraph}~\cite{ibm_langgraph} explicitly models agent workflows as directed graphs of nodes (decisions, tool uses, conditionals), increasing transparency over linear chains. However, these frameworks leave data flow largely implicit---the system manages prompts and memory behind the scenes, making it difficult to audit what each step actually ``sees.''

\textbf{NormCode differs in enforcing explicit data isolation.} While LangGraph provides structural transparency (the graph topology is visible), NormCode provides \textit{data-flow transparency}: every input to every step is explicitly declared. When debugging a failure at step 7, NormCode allows precise inspection of step 7's inputs, not just its position in the workflow graph.

\subsection{Intermediate Representations and Structured Reasoning}

The concept of an intermediate representation (IR) between high-level intent and low-level execution has deep roots in computer science~\cite{wiki_ir}. In compilers, an IR (e.g., LLVM IR) abstracts away machine details, enabling optimization passes independent of source language. A good IR is accurate (captures all essential information) and modular (supports transformation without loss of meaning).

In LLM research, structured reasoning traces serve analogous purposes. \textbf{Chain-of-Thought (CoT)} prompting~\cite{wei2022} encourages models to generate step-by-step reasoning, significantly improving complex problem-solving. \textbf{Tree-of-Thought (ToT)}~\cite{yao2023} generalizes this by exploring branching paths and backtracking. \textbf{Graph-of-Thought (GoT)}~\cite{han2023} allows non-linear exploration of reasoning networks.

NormCode can be viewed as an IR for AI planning that combines these insights: it provides a structured ``language of thought'' that is simultaneously human-authored (like natural CoT), machine-executable (like compiled code), and inspectable (like a computation graph). Recent work on ``Abstractions-of-Thought''~\cite{delorenzo2025} and ``Blueprint First'' approaches~\cite{qiu2025} uses structured IRs to separate functional decomposition from syntax in LLM-based hardware design---a similar philosophy to NormCode's separation of reasoning structure from execution details.

Critically, NormCode's IR supports \textbf{progressive formalization}: plans can start as rough sketches and be iteratively refined. This contrasts with traditional IRs (which demand full formalization upfront) and free-form CoT (which lacks enforceable structure).

\subsection{Semi-Formal Languages and the Formalization Spectrum}

NormCode occupies a deliberate position between natural language and formal specification. Pure natural language is expressive but ambiguous; fully formal languages (PDDL, code) are unambiguous but rigid. Semi-formal languages strike a balance, imposing structure to reduce ambiguity while remaining accessible to humans~\cite{silva_dl}.

In requirements engineering, UML provides semi-formal diagrams that allow automated consistency checks while staying understandable to stakeholders. NormCode follows this philosophy for AI planning: the language is strict enough for reliable compilation and execution, but flexible enough that humans can author plans in near-natural language (\texttt{.ncds} format) and verify them as readable narratives (\texttt{.ncn} format).

The name ``NormCode'' references normative reasoning---the study of obligations, permissions, and prohibitions in deontic logic~\cite{stanford_deontic}. While NormCode does not explicitly encode deontic modalities, the spirit of ``norms'' suggests rules that agents should follow. Recent work evaluates LLMs' consistency in handling normative constraints~\cite{sordoni2025}, and NormCode's design facilitates such evaluation by making every reasoning step explicit and auditable.

\textbf{Why semi-formal for LLMs?} Models excel at natural language but struggle with fully formal syntax---small errors break rigid parsers. A semi-formal format provides structural guidance (reducing ambiguity) without unforgiving exactness, allowing LLMs to generate valid plans more reliably. This pragmatic balance enables the progressive formalization lifecycle that NormCode supports.

\section{The NormCode Language}

NormCode is designed around a single fundamental unit: the \textbf{inference}. An inference is not merely a function call; it is a structured assertion: ``Concept A is obtained by performing Operation B on Inputs C and D.'' This structure enforces the data isolation required for reliable AI planning.

\subsection{Core Syntax: The Inference Structure}

A NormCode plan is a hierarchy of inferences defined through three primary concept markers. The \texttt{<=} marker denotes a \textbf{Functional Concept}, which defines the operation (the ``verb'') and triggers an agent sequence. The functional concept's reference holds a \textbf{norm}---a paradigm or execution strategy that configures how the agent processes the operation. The \texttt{<-} marker denotes a \textbf{Value Concept}, which defines the data (the ``noun'') holding input or output state. The \texttt{<*} marker denotes a \textbf{Context Concept}, which provides in-loop state or carried data across iterations.

A typical inference in the \texttt{.ncds} (NormCode Draft Straightforward) format appears as:

\begin{verbatim}
<- summary
    <= summarize the text
    <- raw document
\end{verbatim}

This is read bottom-up: ``The \texttt{raw document} is used to \texttt{summarize the text}, producing the \texttt{summary}.'' The indentation defines the dependency: the parent (\texttt{summary}) cannot be resolved until its children are complete.

\subsection{The Multi-Format Ecosystem}

NormCode acknowledges that humans, compilers, and reviewers have different needs. The language exists in multiple isomorphic formats that serve distinct purposes throughout the development lifecycle.

The \textbf{.ncds (Draft Straightforward)} format serves as the human authoring format. It uses natural language with minimal structural markers and prioritizes speed and readability. Authors start here when creating new plans, focusing on rough logic and concept structure without worrying about formal syntax details.

The \textbf{.ncd (Draft)} format is the formal intermediate representation generated by the compiler. It resolves all ambiguities by assigning semantic types, fixing value orders with explicit bindings, and generating unique flow indices for every step. This machine-parsable format contains all technical details required for direct execution by the orchestrator.

The \textbf{.ncn (Natural)} format provides a compiler-generated narrative that translates the formal \texttt{.ncd} back into plain English. For example, an inference might render as ``(OUTPUT) The summary (ACTION) is obtained by summarizing the document (INPUT) using the raw document.'' This format allows non-technical domain experts to verify the plan's logic before execution without understanding the formal syntax.

The \textbf{.ncdn (Hybrid)} format combines both \texttt{.ncd} and \texttt{.ncn} views in a single editor interface. This enables developers to see the formal syntax alongside its natural language interpretation, facilitating review and modification.

Additionally, the \textbf{.nci.json} format provides a JSON representation of the inference structure, showing clear flow relationships between concepts. The final \textbf{.concept.json} and \textbf{.inference.json} formats constitute the executable repositories loaded by the orchestrator, containing separated concept definitions and inference working interpretations respectively.

The typical workflow proceeds from authoring in \texttt{.ncds}, through compiler-driven formalization to \texttt{.ncd} with its \texttt{.ncn} companion, then to structured \texttt{.nci.json}, and finally to the separated executable repositories. The compiler handles most transformations automatically, allowing users to focus on writing clear plans.


\subsection{Semantic Concept Types}

NormCode types are semantic, not just structural. They tell the agent \textit{what} the data represents in the world, and each concept type is backed by a \textbf{Reference}---a multi-dimensional tensor that holds its content.

\textbf{Non-Functional Entities} represent the ``nouns'' and ``states'' of reasoning. \textbf{Objects} (\texttt{\{\}}) represent discrete items such as files, queries, or results. \textbf{Propositions} (\texttt{<>}) represent boolean states or facts that can be true or false, often extracted from declarative clauses or conditions. \textbf{Relations} (\texttt{[]}) represent collections or mappings, typically extracted from plurals or groupings. \textbf{Subjects} (\texttt{:S:}) represent active agents responsible for executing logic, holding a ``Body'' of tools and capabilities.

\textbf{Functional Operations} represent the ``verbs'' that transform data or evaluate states. Their references hold a \textbf{norm} that specifies the paradigm or execution strategy, bridging the Subject (agent with tools) and Object (data). \textbf{Imperatives} (\texttt{(\{\})} or \texttt{::()}) represent commands that change state or produce new data, invoking the Imperative Sequence. \textbf{Judgements} (\texttt{<\{\}>}) represent evaluations that return truth values, invoking the Judgement Sequence with an additional truth assertion step.

The critical distinction is that only functional concepts (imperatives and judgements) invoke LLM calls. Everything else is data structure manipulation.

\subsection{Syntactic Operators}

While semantic concepts invoke AI reasoning, syntactic operators manage the plan's data flow deterministically. These operators constitute a tensor algebra for AI planning, reshaping, combining, traversing, and gating references without examining actual content. This makes them free (no LLM calls), deterministic (same input produces same output), and auditable (exact structural changes are traceable).

The \textbf{Assigning} family (\texttt{\$}) controls which data flows forward. The identity operator (\texttt{\$=}) merges two concept names as aliases. The abstraction operator (\texttt{\$\%}) reifies a literal definition as data. The specification operator (\texttt{\$.}) selects the first valid result from candidates. The continuation operator (\texttt{\$+}) appends data along a specified axis. The selection operator (\texttt{\$-}) extracts elements by index, key, or unpacking.

The \textbf{Grouping} family (\texttt{\&}) controls how multiple references combine. The group-in operator (\texttt{\&[\{\}]}) bundles inputs into a labeled dictionary-like structure where items retain their names as keys. The group-across operator (\texttt{\&[\#]}) flattens inputs into a unified list where items lose their source identity.

The \textbf{Timing} family (\texttt{@}) controls execution flow. The conditional operator (\texttt{@:'}) executes only if a condition is true. The negated conditional (\texttt{@:!}) executes only if a condition is false. The sequencing operator (\texttt{@.}) ensures execution occurs after a dependency completes.

The \textbf{Looping} family (\texttt{*}) controls iteration. The iterate operator (\texttt{*.}) traverses a collection element-by-element while carrying state forward across iterations. Loop variables use version markers: \texttt{*1} for the current value in loop 1, \texttt{*-1} for the previous iteration's value, and \texttt{*0} for the initial value before any iteration.

All syntactic operators share a unified modifier system: \texttt{\%>} indicates source (input), \texttt{\%<} indicates target (output), \texttt{\%:} indicates axis (dimension), \texttt{\%\^{}} indicates carry (state), \texttt{\%@} indicates index (position), and \texttt{\%+} indicates creation of a new axis.

\subsection{Plan Addressability}

Every step in a NormCode plan is assigned a unique \textbf{Flow Index} (e.g., \texttt{1.2.3}) based on its position in the hierarchical structure. Each number represents a level in the tree, with deeper nesting producing longer indices. This indexing system enables precise debugging where errors can be localized to exact steps, targeted intervention where execution can be paused at specific points, auditing where inputs for any step can be inspected, and cross-referencing where steps can explicitly depend on others by index.

Flow indices are generated during the Formalization phase and persist through execution, providing stable addresses that survive plan modifications and enabling checkpoint-based resumption.

\subsection{Semi-Formality: Balancing Structure and Flexibility}

NormCode's ``semi-formal'' nature operates in two complementary ways that address the fundamental tension between expressiveness and precision.

First, NormCode supports \textbf{Conceptual Preservation through Progressive Formalization}. Natural language content is allowed within formal structure, enabling users to write \texttt{::(summarize the text)} without immediately defining implementation details. The structure formalizes \textit{how} concepts relate to others (dependencies and data flow) while leaving \textit{content} flexible until execution. Plans can start as rough sketches and be iteratively refined into rigorous logic, supporting exploration before commitment.

Second, NormCode enforces \textbf{Strictness Only Where Necessary}. Syntax is strict precisely to the extent required for compilation: establishing unique concept identities, resolving data flow dependencies, and extracting working interpretations that determine which agent sequence to invoke.

Beyond these requirements, semantic content can remain in natural language. This prevents the brittleness of traditional formal methods where a single syntax error invalidates an entire specification, while providing sufficient structure for reliable orchestration. The name ``NormCode'' references normative reasoning---rules and norms that agents must follow. Each step follows the norm of ``only see what you're explicitly given,'' and adherence to these norms is structurally verifiable.

\section{The Reference System}

While concepts define the \textit{meaning} of data, the \textbf{Reference System} provides the machinery for storing and manipulating it. In NormCode, every piece of data---from a single file path to a collection of user queries---is encapsulated in a \textbf{Reference}, a structure inspired by multi-dimensional tensors.

\subsection{References as Tensors with Named Axes}

Unlike standard lists or dictionaries, a Reference organizes data along named axes (e.g., \texttt{['student', 'assignment']} rather than \texttt{[0, 1]}). This allows operations to be robust to changes in data shape. A collection of documents is not just a list; it is a tensor with a \texttt{document} axis. If we process each document to extract three features, the result is automatically a 2D tensor with \texttt{['document', 'feature']} axes. This explicit structure prevents the ``shape mismatch'' errors common in ad-hoc prompt chains.

\subsection{Perceptual Signs: The ``Lazy Evaluation'' of AI}

Carrying large data payloads (e.g., full text of books) through every step of a plan is inefficient and creates context pollution. NormCode solves this with \textbf{Perceptual Signs}: lightweight pointers that represent data without loading it.

A sign follows the format \texttt{\%\{Norm\}ID(Signifier)}. For example:
\begin{verbatim}
%{file_location}a1b(data/contract.pdf)
\end{verbatim}

This tells the agent: ``There is an object here (ID \texttt{a1b}). To perceive it, use your \texttt{FileSystem} faculty (the \texttt{file\_location} norm) on the path \texttt{data/contract.pdf}.'' The actual data is only loaded (``transmuted'') at the exact moment an inference needs to operate on it (the \textbf{MVP} step). Until then, the system passes around the lightweight sign, ensuring efficiency.

\subsection{Semantic vs. Syntactic Operations on References}

The distinction between semantic and syntactic operations is implemented at the Reference level through a layered algebra.

\textbf{1. Syntactic Operations (Data Plumbing)}: Operations like \textbf{Reshaping} (\texttt{slice}, \texttt{append}) and \textbf{Combining} (\texttt{cross\_product}, \texttt{join}) manipulate the \textit{structure} of the tensor without looking at the content. They are instant, deterministic, and cost zero tokens.

\textbf{2. Semantic Operations (AI Reasoning)}: The \textbf{Cross-Action} (\texttt{cross\_action(functions\_ref, values\_ref)}) is the bridge between structure and meaning, applying an LLM-prepared function to values element-by-element.

By restricting LLMs to \texttt{cross\_action} and handling all other data movement via syntactic operators, NormCode ensures that AI reasoning is applied only where strictly necessary.

\section{Design Philosophy}

NormCode's design is guided by three core principles that address a fundamental tension in AI systems: the need for human oversight in processes that are increasingly automated.

\subsection{Dual-Readability: Bridging Human and Machine}

AI planning systems face a dilemma. Humans think in natural language; machines require unambiguous instructions. Most systems resolve this by forcing humans to write in a machine format (tedious, error-prone) or by letting machines interpret natural language (opaque, unauditable).

NormCode sidesteps this dilemma through its multi-format ecosystem. The \texttt{.ncds} format enables fast, intuitive authoring in natural language for human authors. The \texttt{.ncd} format provides unambiguous, machine-executable representation for the compiler and orchestrator. The \texttt{.ncn} format offers a readable narrative for verification before execution by human reviewers. The \texttt{.ncdn} hybrid format combines both views for developers working in the editor.

The key insight is that these formats are not translations---they are the same plan at different levels of explicitness. An author writes \texttt{.ncds}, the compiler enriches it to \texttt{.ncd} (adding types, bindings, flow indices), and a reviewer reads \texttt{.ncn} to verify the logic. No information is lost; no ambiguity is introduced. This allows domain experts who may not understand formal syntax to audit AI workflows before they run.


\subsection{Progressive Formalization: From Sketch to Structure}

Traditional formal methods demand rigor upfront: you must specify everything before you can execute anything. This is impractical for AI workflows, where the ``right'' structure often emerges through experimentation.

NormCode supports \textbf{Progressive Formalization}---a lifecycle where plans start loose and tighten over time. During the \textit{Exploration Phase}, authors write rough \texttt{.ncds} sketches where concepts can be vague (``process the document somehow'') and structure captures dependencies without implementation details. During the \textit{Refinement Phase}, authors run the plan, observe failures, and tighten specific inferences by adding type constraints, fixing value orderings, and making concepts explicit. During the \textit{Production Phase}, the plan is rigorous with every step auditable and checkpointing reliable.

This lifecycle is enabled by the semi-formal nature of NormCode: the formalism only demands what the compiler needs. Everything else can remain flexible until you choose to lock it down.

\textbf{Intervenability} is a key feature throughout this lifecycle. Because every step has a unique flow index, a user or automated system can pause execution before a specific step, inspect the inputs a step will receive, modify a concept's reference before resuming, or fork a run to explore alternative branches. The Canvas App provides visual tools for all these interventions.

\subsection{Semantic vs. Syntactic Separation: Cost and Reliability Tracing}

Perhaps the most practically important design decision in NormCode is the clean separation between operations that invoke AI reasoning and operations that simply move data around.

\textbf{Semantic operations} include imperatives and judgements, require LLM calls, consume tokens, and are non-deterministic. These operations create new information through reasoning, generation, or evaluation. \textbf{Syntactic operations} include grouping, assigning, timing, and looping, require no LLM calls, are free, and are fully deterministic. These operations reshape existing information through tensor algebra without examining content.

In a typical NormCode plan, the majority of steps are syntactic. They collect inputs, select outputs, iterate over collections, and branch on conditions---all without any AI involvement. Only the ``thinking'' steps (imperatives and judgements) invoke an LLM.

This separation provides three crucial capabilities. \textbf{Cost Visibility} means you know exactly which steps burn tokens, allowing optimization efforts to focus on expensive operations. \textbf{Reliability Mapping} means syntactic steps never fail unexpectedly, so if a plan fails, the cause is localized to a semantic step. \textbf{Auditability} means for any execution, you can generate a report: ``Steps 1.1, 1.3, 2.2 called the LLM; all other steps were deterministic data routing.''

For high-stakes domains (legal, medical, financial), this transparency is often a regulatory requirement. NormCode makes it structural rather than aspirational.

\subsection{When to Use NormCode}

NormCode adds structure, and structure has costs. The framework is not appropriate for every use case.

\textbf{Strong fit scenarios} include multi-step workflows with five or more LLM calls where isolation and debuggability pay off, auditable AI applications in legal, medical, or financial domains where you must prove what each step saw, long-running resumable workflows where built-in checkpointing adds value, and human-AI collaboration where domain experts need to inspect and modify plans.

\textbf{Poor fit scenarios} include quick prototypes with one or two LLM calls where overhead exceeds benefit, simple Q\&A chatbots where direct prompting suffices, and real-time applications where orchestration latency is unacceptable.

The sweet spot is complex, multi-step workflows where you need to know exactly what happened at each step---and where a failure in step 7 should not corrupt the reasoning in step 12. This is precisely the scenario where context pollution in traditional approaches causes the most damage, and where NormCode's enforced data isolation provides the most value.

\section{The Execution Model}

While the NormCode language defines \textit{what} a plan should do, the execution model defines \textit{how} and \textit{when} it happens. This section describes the Orchestrator (the central execution engine), Agent Sequences (execution pipelines for each inference type), Paradigms (declarative configuration of agent behavior), and the Canvas App (a visual debugging environment).

\subsection{The Orchestrator}

The Orchestrator is the runtime engine of NormCode, managing plan execution through dependency tracking, inference scheduling, and state maintenance. Its architecture centers on three core components.

The \textbf{Waitlist} maintains a prioritized queue of all inferences in the plan, sorted by flow index. This defines the structural order of execution and provides a complete manifest of work to be performed.

The \textbf{Blackboard} serves as a real-time state tracker where every concept and inference has a status: \texttt{pending} (not yet started), \texttt{in\_progress} (currently executing), \texttt{completed} (finished successfully), or \texttt{skipped} (bypassed due to timing conditions). The Blackboard is the single source of truth for execution state.

The \textbf{Repositories} consist of the Concept Repository (storing References for each concept) and the Inference Repository (storing configuration for each inference including which sequence to run and which paradigm to use).

The Orchestrator operates in cycles. In each cycle, it scans the Waitlist for pending inferences whose dependencies are met (all input concepts are completed), executes ready inferences by invoking the appropriate Agent Sequence, and updates the Blackboard and Concept Repository with results. This dependency-driven scheduling ensures that inferences run only when their inputs are ready and that failures in one branch do not corrupt unrelated branches.


\subsection{Persistence and Checkpointing}

For long-running or resumable workflows, the Orchestrator provides a robust checkpointing system backed by SQLite. Each execution is assigned a unique Run ID, and the full state (Blackboard, References, Workspace) is saved at the end of each cycle as a snapshot.

The system supports multiple resume modes for handling runs that restart after code changes. The \texttt{PATCH} mode (default) performs a smart merge, re-running inferences whose definitions have changed while keeping valid cached states. The \texttt{OVERWRITE} mode trusts the checkpoint entirely without reconciliation. The \texttt{FILL\_GAPS} mode only populates missing data from the current repository defaults.

Fork functionality allows loading a past state while starting a new run history, useful for branching experiments, retrying with modified logic, or A/B testing different configurations.

\subsection{Agents and the AgentFrame}

In NormCode, an \textbf{Agent} is represented by the Subject Concept (\texttt{:S:}) and constitutes a concrete container of capability. When a functional concept executes, it does so within an Agent's context, accessing that agent's specific tools and configurations.

The \textbf{AgentFrame} class realizes agents in the implementation, containing three essential components. The \textbf{Body} provides the agent's ``toolbox''---a registry of available tools such as LLM clients, file system access, and Python executors. The \textbf{Sequences} define the logic pipelines the agent can run, including imperative, judgement, grouping, assigning, timing, and looping. The \textbf{Mode} specifies the interpretation style, such as composition mode for paradigm-driven execution.

This design enables multi-agent planning where different Subjects (e.g., \texttt{:coder\_agent:} and \texttt{:reviewer\_agent:}) can have different tool bodies and capabilities even within the same plan. Inferences can be assigned to specific agents via pattern rules matching flow indices, concept names, or sequence types, or through explicit assignment of specific inferences to agent identifiers.

\subsection{Agent Sequences: The Execution Pipelines}

Each inference type triggers a specific Agent Sequence---a standardized pipeline of steps that follows a cognitive cycle of Perception, Actuation, and (for judgements) Assertion.

\textbf{Semantic Sequences} invoke LLM calls to create new information. The \textbf{Imperative Sequence} follows seven steps: IWI (Input Working Interpretation) parses syntax and determines the paradigm; IR (Input Reference) retrieves references for all input concepts; MFP (Model Function Perception) transforms the functional definition into an executable function using paradigms; MVP (Memory Value Perception) retrieves and transmutes perceptual signs into actual data; TVA (Tool Value Actuation) applies the prepared function to perceived values; OR (Output Reference) creates or updates the concept's reference with results; and OWI (Output Working Interpretation) updates execution state and logs completion.

The \textbf{Judgement Sequence} extends the imperative sequence with an additional step: TIA (Tool Inference Assertion) checks if TVA results satisfy truth conditions by applying a quantifier (all, any, exists) and evaluating a condition, collapsing the result into a boolean reference.

\textbf{Syntactic Sequences} perform deterministic data manipulation without LLM involvement. The \textbf{Assigning Sequence} (steps IWI-IR-AR-OR-OWI) handles variable assignment and selection through specification (selecting first valid), continuation (appending along an axis), or derelation (structural extraction). The \textbf{Grouping Sequence} (IWI-IR-GR-OR-OWI) collects and combines data using and-in (labeled bundling) or or-across (flat concatenation). The \textbf{Timing Sequence} (IWI-T-OWI) controls conditional execution by checking progress conditions, if-conditions, or negated if-conditions against the Blackboard. The \textbf{Looping Sequence} (IWI-IR-GR-LR-OR-OWI) manages iteration through retrieving next elements, storing iteration results in workspace, and combining all looped elements upon completion.

These syntactic sequences are instant, free, and fully deterministic, handling all data plumbing so that semantic operations receive exactly what they need.

\subsection{Paradigms: Configuring Agent Behavior}

A \textbf{Paradigm} is a declarative JSON specification that configures how an agent executes a semantic inference, bridging the gap between abstract intent (``summarize this document'') and concrete execution (which prompt template, which LLM, which output format).

Paradigms separate configuration into two phases following a vertical/horizontal split. \textbf{Vertical Steps} occur during the MFP phase at construction time, resolving tool affordances and producing callable functions. These steps load LLM clients, configure model settings, and prepare generation functions from the agent's body. \textbf{Horizontal Steps} occur during the TVA phase at runtime, passing actual data values through the pre-configured functions. These steps receive prompts from value concepts, execute LLM calls, extract results, and optionally save outputs.

Paradigm files follow a self-documenting naming convention: \texttt{h\_} prefixes indicate horizontal inputs, \texttt{v\_} prefixes indicate vertical inputs, \texttt{c\_} prefixes indicate composition steps, and \texttt{o\_} prefixes indicate output formats. For example, \texttt{h\_PromptTemplate\_SavePath-c\_GenerateThinkJson-
Extract-Save-o\_FileLocation.json} describes a paradigm that takes a prompt template and save path as horizontal inputs, composes LLM generation with JSON extraction and file saving, and outputs a file location pointer.

Mathematically, paradigm execution can be formalized as: let $\mathcal{S}$ be the Agent's state, $V_{spec}$ be the vertical specification, $H_{plan}$ be the horizontal plan, and $\mathcal{V}$ be runtime values. The output is computed as:

\begin{equation}
\mathcal{O} = [F_C(F_V(\mathcal{S}, V_{spec}), H_{plan})](\mathcal{V})
\end{equation}

where $F_V$ is the vertical function (MFP phase) producing tool handles $\mathcal{T}$, $F_C$ is the composition function compiling $\mathcal{T}$ and $H_{plan}$ into function $\Phi$, and $\Phi(\mathcal{V})$ is the horizontal execution (TVA phase) producing output $\mathcal{O}$. This clean separation allows paradigms to be reused across different inferences and agents.

\subsection{The Canvas App: Visual Debugging}

The NormCode Graph Canvas App provides a visual, interactive environment for executing, debugging, and auditing NormCode plans. Built on React Flow for graph visualization and FastAPI for the backend, it transforms NormCode from ``run and hope'' to ``observe and control.''

The Canvas App enables users to visualize the entire inference graph before execution, with semantic functions shown in purple and semantic values in blue, while syntactic operations appear in gray. Users can watch execution progress in real-time through WebSocket-streamed events, with node status indicators showing pending (gray), running (blue pulsing), completed (green), failed (red), or skipped (striped) states.

The debugging capabilities include setting breakpoints on specific flow indices to pause execution at critical points, stepping through inferences one at a time, and inspecting tensor data at any node through a TensorInspector component that handles N-dimensional viewing with table, list, or JSON modes.

The multi-agent configuration panel allows registering multiple agents with different LLM models (such as qwen-plus, gpt-4o, or claude-3), tool settings, and paradigm directories. Inferences can be mapped to specific agents through pattern rules or explicit assignments, with a default agent handling unmatched inferences.

Tool call monitoring captures all LLM calls with prompts and responses, file system operations, Python script executions, and user input requests in real-time. The checkpoint panel enables resuming from or forking past execution states.

\begin{figure*}[t]
\centering
\includegraphics[width=\textwidth]{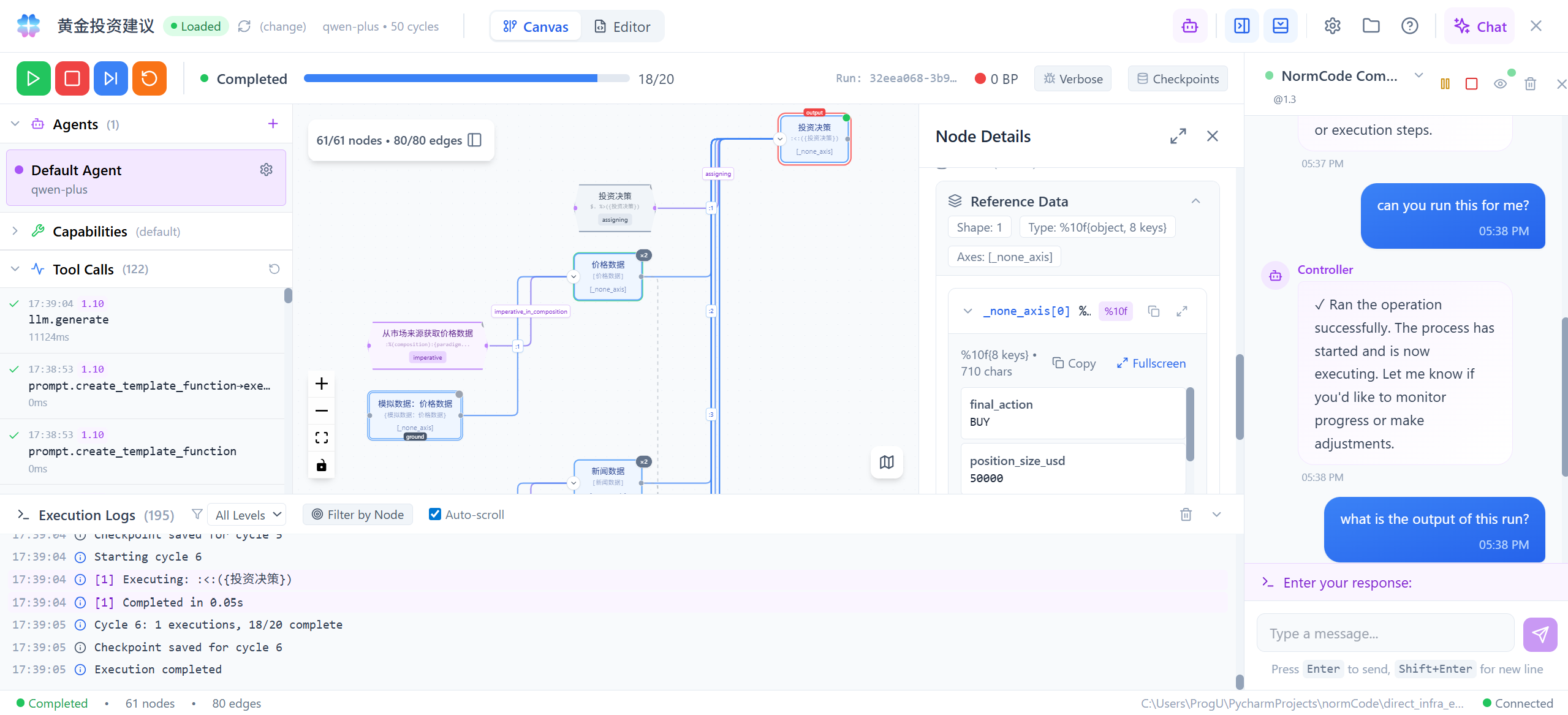}
\caption{The NormCode Canvas App visual debugging environment. The graph visualization (center) shows the inference flow with semantic operations in purple/blue and syntactic operations in gray. The left panel provides agent configuration and tool call monitoring. The right panel displays node details with tensor inspection, reference data, and execution state. Progress indicators show real-time execution status with 18/20 nodes completed.}
\label{fig:canvas_app}
\end{figure*}

\subsection{Command-Line Interface}

For headless execution, the CLI orchestrator provides essential commands: \texttt{run} starts a fresh execution with specified concept and inference repositories; \texttt{resume} continues from the latest checkpoint with an optional Run ID; \texttt{fork} branches from a past state with a new run ID; and \texttt{list-runs} shows all tracked executions in the database.

\section{The Compiler Ecosystem}

The NormCode compiler transforms human-authored plans into executable artifacts through a multi-stage pipeline. Each stage progressively adds rigor while preserving opportunities for human review, following a progressive formalization philosophy where each phase answers a specific question while preserving the semantic intent of previous phases.

\subsection{The Four-Phase Pipeline}

The current implementation follows a four-phase pipeline that transforms natural language intent into executable JSON repositories. The compiler handles most transformations automatically, with optional manual review between phases.


\textbf{Phase 1: Derivation} takes natural language input and produces \texttt{.ncds} (draft straightforward) output. This phase answers the question ``WHAT are we trying to do?'' by identifying concepts (data entities), identifying operations (actions to perform), extracting dependencies (which concepts feed into which operations), and creating hierarchical tree structure. The derivation phase may use LLM assistance for decomposing complex natural language instructions into structured concepts.

\textbf{Phase 2: Formalization} takes \texttt{.ncds} input and produces \texttt{.ncd} (formal) with an \texttt{.ncn} companion. This phase answers ``IN WHAT ORDER and WHICH SEQUENCE?'' by assigning unique flow indices (e.g., \texttt{1.2.3}) to every step, determining sequence types for each inference (semantic types like imperative and judgement, or syntactic types like assigning, grouping, timing, and looping), resolving concept identity and value bindings using markers like \texttt{<:\{1\}>}, and adding metadata comments for sequence markers and flow indices.

\textbf{Phase 3: Post-Formalization} takes the formal \texttt{.ncd} and produces an enriched \texttt{.ncd} with full annotations. This phase answers ``HOW and WITH WHAT RESOURCES?'' through three sub-phases. The \textit{Re-composition} sub-phase maps abstract intent to the normative context of execution by adding paradigm IDs, perception norms for vertical and horizontal inputs, body faculty specifications, and output structure hints. The \textit{Provision} sub-phase fills in concrete resources by specifying file locations for ground data and paths to prompts or scripts. The \textit{Syntax Re-confirmation} sub-phase confirms reference structure by declaring axis names, tensor shapes, element types, and filter documentation for tensor coherence.

\textbf{Phase 4: Activation} takes the enriched \texttt{.ncd} and produces the executable \texttt{.concept.json} and \texttt{.inference.json} repositories. This phase answers ``WHAT DOES THE ORCHESTRATOR NEED?'' by extracting concept definitions with types, axes, and ground values; extracting inference definitions with function concepts, value concepts, and context concepts; and generating \texttt{working\_interpretation} dictionaries containing exactly what each sequence's IWI step expects.

Each phase includes an opportunity for manual review, enabling human intervention before the next stage proceeds. This supports progressive formalization where the plan tightens incrementally rather than requiring complete specification upfront.

\subsection{Derivation: Natural Language to Structure}

The derivation phase transforms unstructured natural language into hierarchical inference structure. For complex instructions, this may use a recursive decomposition algorithm where the entire instruction is wrapped as a top-level concept with a source text annotation, un-decomposed concepts are iteratively identified and questions formulated about their source text, question types are classified and mapped to appropriate NormCode operators, and decomposition continues until no source annotations remain.

This process can be guided by a taxonomy of question types that map to specific NormCode structures. Methodology declarations produce functional concepts with action semantics. Conditional dependencies produce timing operators. Iterative requirements produce looping operators. Collection requirements produce grouping operators.

\subsection{Formalization: Adding Structural Rigor}

Once structure is established, formalization adds precision required for execution. Serialization reframes each step as an output-effect relationship, making explicit that a concept is produced by applying a function to inputs. Redirection links abstract references to concrete implementations through annotations specifying prompts, scripts, and file locations. Flow index generation assigns unique hierarchical addresses to every line, enabling precise identification and cross-referencing.

The result is an \texttt{.ncd} file where every inference has explicit inputs, explicit outputs, and a unique identity. The companion \texttt{.ncn} file translates this structure back into natural language for verification by domain experts who may not understand the formal syntax.

\subsection{Post-Formalization: Enriching for Execution}

Post-formalization prepares the plan for execution by adding configuration and grounding that the orchestrator requires. This phase operates within a normative context defined by the agent's Body (available tools), Paradigms (action norms stored in configuration files), and PerceptionRouter (perception norms for transmuting perceptual signs).

Annotations added during post-formalization include paradigm specifications (\texttt{\%\{norm\_input\}}) identifying which paradigm file to load, perception norm specifications for vertical and horizontal inputs, body faculty specifications identifying which tool to invoke, and reference structure declarations specifying axes, shapes, and element types.

\subsection{Activation: Generating Executable Repositories}

The final compilation stage transforms the enriched \texttt{.ncd} into two JSON repositories that the Orchestrator can load and execute.

The \textbf{Concept Repository} (\texttt{concept\_repo.json}) stores static definitions of all data entities. Each entry contains the concept name with semantic type markers, the type indicator, flags for ground concepts (pre-initialized with data) and final concepts (root outputs), initial reference data as perceptual signs where applicable, and reference axis names defining tensor structure.

The \textbf{Inference Repository} (\texttt{inference\_repo.json}) stores operational definitions. Each entry contains flow information with the unique index, the inference sequence type determining which pipeline executes, the concept to infer identifying the output, the function concept defining the operation, value concepts listing inputs, context concepts listing loop state, and crucially the \texttt{working\_interpretation} dictionary.

The \texttt{working\_interpretation} is the critical output of activation, containing exactly what each sequence's IWI step expects. For imperative sequences, this includes paradigm identification, value ordering, and optional value selectors for decomposing grouped concepts. For judgement sequences, it adds assertion conditions with quantifiers and truth values. For assigning sequences, it specifies the marker type, source, destination, and any axis information. For grouping sequences, it specifies the grouping mode, axes to collapse, and any protected axes. For timing sequences, it specifies the condition type and referenced concept. For looping sequences, it specifies the loop base, current element, group base, in-loop concepts, and concepts to infer.

\subsection{Translation: Generating Human-Readable Views}

For human verification, the system generates Natural Language NormCode (\texttt{.ncn}) files that strip formal markers and present plans as readable prose. A formal inference such as:

\begin{verbatim}
<- {Phase 1: Confirmation of Instruction}
    <= &[#] %>[{step 1.1}, {step 1.2}] %+(step)
    <- {step 1.1: Instruction Distillation}
    <- {step 1.2: Context Registration}
\end{verbatim}

translates to natural language as:

\begin{verbatim}
(OUTPUT) Phase 1: Confirmation of Instruction
    (ACTION) is obtained by collecting 
    the following steps together.
    (VALUE) The first step is Instruction 
    Distillation.
    (VALUE) The second step is Context 
    Registration.
\end{verbatim}

This enables domain experts to verify plan logic before execution without understanding formal syntax. The hybrid \texttt{.ncdn} format presents both views side-by-side for developers who need to see the mapping between formal and natural representations.

\subsection{Round-Trip Capability}

The compilation pipeline supports bidirectional transformation, enabling JSON repositories to be decompiled back to \texttt{.ncd} format. This round-trip capability supports debugging by inspecting compiled output in readable format, version control by storing plans as text rather than JSON, and modification by editing compiled plans and re-activating them.

\subsection{Compilation Guarantees and Limitations}

Compilation ensures syntactic validity (every \texttt{.ncd} is parseable), flow consistency (indices are unique and hierarchical), sequence completeness (every inference has a valid sequence type), reference structure (all concepts have declared axes and types), resource grounding (all perceptual signs are linked), and working interpretation completeness (every inference has full configuration).

However, compilation does not ensure semantic correctness (the plan might not achieve intended goals), runtime success (LLM calls might fail or resources might be missing), logical soundness (dependency cycles are detected by the orchestrator, not the compiler), or optimal performance (some plans might be inefficient).

The distinction is important: compilation validates structure, not intent.

\section{Evaluation}

We validate NormCode through two complementary demonstrations: a \textbf{self-hosted compiler pipeline} demonstrating expressive completeness and a \textbf{base-X addition algorithm} demonstrating correctness and debuggability. Additionally, we describe the working implementation of the Canvas App as evidence of practical viability.

\subsection{Case Study 1: The Self-Hosted Compiler Pipeline}

The most significant validation of NormCode is that its own compiler pipeline can be expressed and executed as a NormCode plan.

The task was to transform a high-level natural language instruction into an executable NormCode plan, complete with context distribution and prompt generation. The plan consisted of approximately 50 inferences organized across the four compilation phases: Derivation to extract structure from natural language, Formalization to add flow indices and sequence types, Post-Formalization to enrich with paradigms and resources, and Activation to generate executable JSON repositories.

The orchestrator successfully executed this plan, producing the concept and inference repositories along with all intermediate artifacts. Checkpointing allowed the compilation to be paused and resumed across multiple sessions, and the Canvas App enabled visual inspection of intermediate states.

This self-hosting demonstrates three key properties. First, \textbf{Expressive Completeness}: NormCode can represent complex, multi-phase workflows with loops, conditionals, and human-in-the-loop steps. Second, \textbf{Practical Viability}: The orchestrator, checkpointing, paradigm systems, and visual debugger work together to execute real plans. Third, \textbf{Recursive Validation}: Any improvement to NormCode can be tested by re-running its own compilation pipeline.

\subsection{Case Study 2: Base-X Addition Algorithm}

To validate the correctness of NormCode-orchestrated execution, we implemented a base-X addition algorithm. This serves as a controlled benchmark where correctness can be unambiguously verified against known mathematical results.

The task was adding two arbitrary-base numbers (base-10, base-12, etc.) digit-by-digit, handling carry-over correctly across arbitrary digit lengths. The NormCode plan decomposed into approximately 25 inferences organized around three nested loops: an outer loop over number pairs in the collection, an inner loop extracting unit-place digits from each number, and a parallel loop generating the shifted number pairs for the next iteration.

The Concept Repository contained over 50 concepts covering ground data such as the initial number pairs, intermediate states such as digit sums and carry-over numbers and remainders, and control predicates such as whether all numbers are zero and whether the carry-over is zero.

Key mechanisms demonstrated include \textbf{Loop Quantification} where the outer loop iterates until termination while carrying carry-over state between iterations; \textbf{Conditional Branching} with nested timing operators controlling when to stop appending based on whether numbers are exhausted and no carry remains; \textbf{Grouping} where digits from both numbers are collected into a single collection for summation; and \textbf{Timing Dependencies} ensuring quotient and remainder are computed only after the digit sum is available.

The orchestrator achieved 100\% accuracy on the addition task across test suites of varying digit lengths, validated up to 150-digit numbers. Because NormCode ultimately generates deterministic code via paradigms that invoke Python execution, the final output is deterministic. The role of NormCode is to structure the derivation of that code, ensuring each step receives exactly the correct context and the overall logic remains auditable.

\subsection{Implementation Evidence: The Canvas App}

Beyond the case studies, the working Canvas App implementation provides evidence of practical viability. The application was developed using React 18 with TypeScript for the frontend, Vite as the build tool, React Flow for graph visualization, Zustand for state management, and TailwindCSS for styling. The backend uses FastAPI with Python 3.11, WebSockets for real-time event streaming, and Pydantic for data validation.

The Canvas App successfully demonstrates real-time graph visualization where inference graphs are rendered with appropriate node types and status indicators; WebSocket event streaming where execution events flow in real-time from backend to frontend; breakpoint debugging where execution pauses at specified flow indices; tensor inspection where N-dimensional reference data is viewable through multiple modes; multi-agent configuration where different agents can be registered with different LLM models and tool configurations; and checkpoint management where past states can be resumed or forked.


\subsection{Qualitative Observations}

While formal experiments with statistical analysis are planned for future work, qualitative observations from using NormCode in practice support several claims.

Regarding \textbf{Debuggability}, flow indices enable precise localization of failures. When an inference fails, the log identifies exactly which step (e.g., ``Step 1.3.2'') failed and what inputs it received. The Canvas App makes this visual, showing the failed node in red with its input connections highlighted.

Regarding \textbf{Token Efficiency}, perceptual signs reduce token cost by passing lightweight pointers instead of full data. Only during the MVP step, when an inference actually needs to operate on data, are the signs transmuted into actual content. Syntactic operations are completely free, incurring no LLM costs regardless of the data volume they manipulate.

Regarding \textbf{Resumability}, SQLite-backed checkpointing allows runs to be paused and resumed without loss of state. This is particularly valuable for long-running compilations or iterative development where you want to inspect intermediate states before continuing.

Regarding \textbf{Auditability}, every inference's inputs and outputs are logged with their full tensor structure. Post-hoc analysis can reconstruct exactly what data each step saw and produced, enabling compliance reporting and debugging of unexpected results.

Comparative evaluation against direct prompting, LangChain, and other frameworks on established benchmarks is planned for future work.

\section{Discussion}

\subsection{When to Use NormCode}

NormCode is designed for scenarios where explicit auditability and data isolation justify the overhead of structured planning. The framework provides strong value in several contexts.

For \textbf{high-stakes decisions} in legal reasoning, medical diagnosis, and financial analysis, traceability is often required by regulation or professional standards. NormCode's audit trail satisfies requirements like those in the EU AI Act for documenting how AI systems reach conclusions.

For \textbf{complex multi-step reasoning} involving five or more LLM calls, debugging implicit data flow becomes prohibitive. When a failure occurs at step 7, knowing exactly what step 7 received---not just its position in the workflow---is essential for diagnosis.

For \textbf{long-running workflows} that span hours or days, checkpointing and resumption prevent loss of progress. The ability to fork from checkpoints enables experimentation without risking the main execution path.

For \textbf{human-AI collaboration} where domain experts need to understand and approve AI plans before execution, the multi-format ecosystem allows technical teams to work in \texttt{.ncd} while domain experts review \texttt{.ncn} narratives.

The framework provides poor fit for simple Q\&A applications with one or two LLM calls where direct prompting suffices, for rapid prototyping where formalization overhead exceeds exploration value, and for real-time applications where orchestration latency is unacceptable.

The sweet spot is workflows where reliability and transparency outweigh the cost of explicit structure---precisely the scenarios where context pollution in traditional approaches causes failures that are difficult to diagnose.

\subsection{Limitations and Tradeoffs}

\textbf{Syntax density.} The \texttt{.ncd} format's markers (\texttt{<=}, \texttt{<-}, \texttt{<\$(\{...\})\%>}, \texttt{<:\{1\}>}) create visual clutter that can be challenging to parse. While the \texttt{.ncds} format mitigates this for authoring and the \texttt{.ncn} format provides readable output, debugging compiled plans requires familiarity with dense notation. The Canvas App's visual representation helps by showing structure graphically rather than textually, but IDE plugins with syntax highlighting and folding would further improve usability.

\textbf{Verbosity.} Simple operations expand to multiple lines. A three-step workflow that could be a 10-line Python script becomes a 30-line NormCode plan with explicit concept markers, flow indices, and sequence annotations. This verbosity is the price of explicit data flow. However, for complex workflows, the marginal cost decreases---a 20-step workflow is not 10 times more verbose than a 2-step workflow, and the debugging benefits compound with complexity.

\textbf{Brittleness of manual editing.} The \texttt{.ncd} format is indentation-sensitive and tightly coupled to flow indices. Manual edits risk breaking dependencies in ways that are difficult to diagnose. The compiler should be the primary way to modify plans, but this creates a barrier for quick fixes. The \texttt{.ncdn} editor format and the Canvas App's editing capabilities partially address this by validating changes in real-time.

\textbf{Tooling dependency.} NormCode requires the orchestrator, compiler, reference system, and ideally the Canvas App for effective use. This ``batteries included'' approach limits portability compared to plain Python or LangChain code that can run anywhere with the base libraries installed. However, the integrated toolchain is necessary for the guarantees NormCode provides---isolation, checkpointing, and auditability cannot be retrofitted onto simpler frameworks.

\textbf{Compiler maturity.} The natural language to \texttt{.ncd} derivation phase relies on carefully designed prompts and is sensitive to instruction complexity. Robust error recovery, automated prompt generation, and handling of edge cases remain open challenges. The formalization, post-formalization, and activation phases are more stable since they operate on structured input.

\subsection{Broader Implications}

NormCode demonstrates that \textbf{structured intermediate representations can bridge human intuition and machine rigor in AI workflows}. The multi-format ecosystem (author in \texttt{.ncds}, execute via \texttt{.ncd}, verify in \texttt{.ncn}) suggests a general pattern for transparent AI systems: provide multiple views of the same underlying logic, each optimized for a different stakeholder.

The semantic/syntactic separation has implications beyond NormCode. As LLM-based systems move into production, operators will need precise cost attribution (``which steps burned tokens?'') and reliability mapping (``which failures were probabilistic vs. deterministic?''). NormCode's architectural separation makes these questions answerable by construction rather than requiring post-hoc log analysis.

The Canvas App demonstrates that visual debugging tools for AI workflows are feasible and valuable. Real-time graph visualization, breakpoint debugging, and tensor inspection are capabilities that other frameworks could adopt. The separation between execution logic (orchestrator) and visualization (Canvas App) shows how these concerns can be decoupled.

The checkpoint and fork capabilities point toward a broader vision of AI workflow development as an iterative, experimental process. Rather than writing a plan and hoping it works, developers can run partially, inspect state, fork to try alternatives, and merge successful branches---a workflow more similar to software development with version control than traditional scripting.

\section{Future Work}

While NormCode provides a working implementation with visual debugging, checkpointing, and multi-agent support, several directions remain for future development.

\textbf{Compiler robustness.} The natural language to \texttt{.ncd} derivation phase remains the most fragile component, relying on carefully designed prompts that are sensitive to instruction complexity. Future work should explore fine-tuned models specifically trained on NormCode generation using human-authored plans as training data, iterative refinement loops with automated validation that detect and correct common decomposition errors, and hybrid approaches where humans sketch high-level structure and LLMs fill in detailed specifications.

\textbf{Multi-agent coordination.} The current implementation supports multiple Subjects with different tool bodies and allows mapping inferences to specific agents, but real multi-agent coordination remains unexplored. Future work could address negotiation protocols where agents with different capabilities agree on task allocation, delegation mechanisms where agents can assign sub-tasks to specialized agents, and conflict resolution when multiple agents produce contradictory results. NormCode's isolation guarantees provide a foundation for safe agent-to-agent communication since each agent's view is explicitly defined.

\textbf{Domain-specific extensions.} High-stakes domains may require specialized semantic types such as \texttt{\{legal precedent\}}, \texttt{\{patient record\}}, or \texttt{\{financial instrument\}} that carry domain-specific validation rules and constraints. Domain-specific verification could enforce rules like ``all medical diagnoses must cite evidence'' or ``all financial recommendations must include risk disclosures.'' Compliance reporting could automatically generate audit trails formatted for regulatory requirements such as EU AI Act documentation.

\textbf{Empirical evaluation.} Rigorous comparative studies would strengthen NormCode's claims. Benchmarking against baseline approaches including direct prompting, LangChain pipelines, and AutoGPT on established datasets like HumanEval or GAIA would quantify cost-accuracy tradeoffs. User studies with domain experts (lawyers, clinicians, analysts) would assess the practical value of the multi-format ecosystem and visual debugging. Longitudinal studies of production deployments would reveal real-world failure modes and optimization opportunities.

\textbf{Performance optimization.} While syntactic operations are free, semantic operations incur token costs and latency. Future work could explore batching similar LLM calls within loops to reduce API overhead, caching and memoization of paradigm outputs for repeated patterns, and parallel execution of independent semantic operations when dependency analysis permits.

\textbf{Integration with external systems.} Production deployments require integration with existing infrastructure. This includes connecting to enterprise LLM endpoints with authentication and rate limiting, integrating with logging and monitoring systems for observability, and supporting deployment patterns like containerization and serverless execution.

\section{Conclusion}

As large language models enable increasingly sophisticated multi-step reasoning, the problem of context pollution threatens to undermine their reliability. When information accumulates implicitly across reasoning steps, models hallucinate, confuse intermediate outputs, and lose track of constraints---making complex workflows paradoxically more brittle than simple ones.

NormCode addresses this through \textbf{enforced data isolation}: each inference operates in a sealed environment with only explicitly passed inputs. This is not a convention but a language-level constraint that the compiler and orchestrator enforce. Combined with a clean semantic/syntactic separation (probabilistic LLM reasoning vs. deterministic data flow), NormCode makes AI workflows auditable by construction.

The multi-format ecosystem (\texttt{.ncds} for authoring, \texttt{.ncd} for execution, \texttt{.ncn} for verification, \texttt{.ncdn} for hybrid editing) embodies a broader principle: transparent AI systems should provide multiple views of the same underlying logic, each optimized for different stakeholders. Domain experts can verify plans in natural language; developers can debug in formal syntax; auditors can trace decisions with precision; and the orchestrator can execute with formal rigor.

The four-phase compilation pipeline (Derivation, Formalization, Post-Formalization, Activation) enables progressive formalization where plans tighten incrementally from rough sketches to production-ready specifications. Each phase answers specific questions while preserving semantic intent, and manual review opportunities allow human intervention at critical junctures.

The Canvas App provides visual debugging capabilities that transform NormCode from ``run and hope'' to ``observe and control.'' Real-time graph visualization, breakpoint debugging, tensor inspection, and multi-agent configuration make the execution process transparent and manipulable.

We validate the approach through self-hosted execution (NormCode's compiler runs as a NormCode plan) and algorithmic correctness (100\% accuracy on base-X addition). These demonstrations confirm that structured intermediate representations can bridge human intuition and machine execution without sacrificing either.

NormCode is not appropriate for every use case---simple tasks do not justify the overhead, and real-time applications may not tolerate orchestration latency. But for high-stakes domains where failures have consequences (legal, medical, financial), the ability to answer ``What did step 7 actually see?'' is not optional. By making context isolation structural rather than aspirational, NormCode provides a foundation for AI systems that are reliable, transparent, and auditable---requirements that will only intensify as LLMs move from research prototypes into production systems.

\bibliography{references}

@techreport{chang2025,
  author       = {Chang, E. and others},
  title        = {{SagaLLM}: Context Management, Validation, and Transaction Guarantees for Multi-Agent {LLM} Planning},
  institution  = {Stanford University},
  year         = {2025}
}

@misc{breunig2025,
  author       = {Breunig, D.},
  title        = {How Long Contexts Fail},
  year         = {2025},
  howpublished = {dbreunig.com Blog},
  note         = {Accessed: 2025}
}

@misc{patel2025,
  author       = {Patel, M.},
  title        = {Why {LangChain} Fails in Production: 7 Hidden Problems},
  year         = {2025},
  howpublished = {LinkedIn post},
  note         = {Accessed: 2025}
}

@misc{ruan2024,
  author       = {Ruan, J.},
  title        = {Context Engineering in {LLM}-Based Agents},
  year         = {2024},
  howpublished = {Medium},
  note         = {Accessed: 2025}
}

@inproceedings{wu2025,
  author       = {Wu, Y. and others},
  title        = {{IsolateGPT}: An Execution Isolation Architecture for {LLM}-Based Agentic Systems},
  booktitle    = {NDSS Symposium 2025},
  year         = {2025}
}

@misc{wand2025,
  author       = {Wand AI Research and Fatemi, M.},
  title        = {Compounding Error Effect in Large Language Models: A Growing Challenge},
  year         = {2025},
  howpublished = {Wand AI Blog},
  note         = {Accessed: 2025}
}

@misc{ibm2025,
  author       = {Winland, V. and Noble, J.},
  title        = {What is {LLM} Orchestration?},
  year         = {2025},
  howpublished = {IBM Think Blog},
  note         = {Accessed: 2025}
}

@article{delorenzo2025,
  author       = {DeLorenzo, M. and others},
  title        = {Abstractions-of-Thought: Intermediate Representations for {LLM} Reasoning in Hardware Design},
  journal      = {arXiv preprint arXiv:2505.15873},
  year         = {2025}
}

@article{qiu2025,
  author       = {Qiu, L. and others},
  title        = {Blueprint First, Model Second: A Framework for Deterministic {LLM} Workflow},
  journal      = {arXiv preprint arXiv:2508.02721},
  year         = {2025}
}

@misc{sai2025,
  author       = {Sai, P.},
  title        = {Evaluating {AI} Transparency: What Do Users Really Need to Understand?},
  year         = {2025},
  howpublished = {AI/UI Medium Publication},
  note         = {Accessed: 2025}
}

@misc{saga2025,
  author       = {Modarressi, A. and Xiao, G. and others},
  title        = {{SagaLLM} Reference},
  year         = {2024},
  note         = {Cited in Chang (2025)}
}

@inproceedings{wang2024,
  author       = {Wang, P. and others},
  title        = {{AutoGen}: Enabling Next-Gen {LLM} Applications via Multi-Agent Conversation},
  booktitle    = {COLING 2024},
  year         = {2024}
}

@inproceedings{erol1994,
  author       = {Erol, K. and Hendler, J. and Nau, D. S.},
  title        = {{HTN} planning: Complexity and expressivity},
  booktitle    = {Proceedings of the 12th National Conference on Artificial Intelligence (AAAI-94)},
  pages        = {1123--1128},
  year         = {1994}
}

@article{yao2022,
  author       = {Yao, S. and Zhao, J. and Yu, D. and Narasimhan, K. and Zhao, D. and Cao, Y.},
  title        = {{ReAct}: Synergizing reasoning and acting in language models},
  journal      = {arXiv preprint arXiv:2210.03629},
  year         = {2022}
}

@article{shinn2023,
  author       = {Shinn, N. and Labash, A. and Liu, E. and Prystawski, B. and Park, C. H. and Raffel, C.},
  title        = {Reflexion: Language agents with verbal reinforcement learning},
  journal      = {arXiv preprint arXiv:2303.11366},
  year         = {2023}
}

@misc{ibm_langgraph,
  author       = {IBM},
  title        = {What is {LangGraph}?},
  year         = {2025},
  howpublished = {IBM Think},
  note         = {Retrieved December 2025}
}

@misc{wiki_ir,
  author       = {Wikipedia contributors},
  title        = {Intermediate representation},
  howpublished = {Wikipedia},
  note         = {Accessed: 2025}
}

@inproceedings{wei2022,
  author       = {Wei, J. and Wang, X. and Schuurmans, D. and Bosma, M. and Ichter, B. and Xia, F. and others and Le, Q.},
  title        = {Chain-of-thought prompting elicits reasoning in large language models},
  booktitle    = {ICLR 2023},
  year         = {2022}
}

@article{yao2023,
  author       = {Yao, S. and Zhao, J. and Yu, D. and Narasimhan, K. and Zhao, D. and Cao, Y.},
  title        = {Tree of Thoughts: Deliberate problem solving with large language models},
  journal      = {arXiv preprint arXiv:2305.10601},
  year         = {2023}
}

@article{han2023,
  author       = {Han, X. and Zhao, D.},
  title        = {Beyond chain-of-thought: Effective graph-of-thought reasoning in language models},
  journal      = {arXiv preprint arXiv:2305.16582},
  year         = {2023}
}

@incollection{silva_dl,
  author       = {Silva, A. R.},
  title        = {A sea of description languages},
  booktitle    = {Software Engineering Companion: Requirements Engineering},
  year         = {2025},
  note         = {Accessed: 2025}
}

@incollection{stanford_deontic,
  author       = {{Stanford Encyclopedia of Philosophy}},
  editor       = {Zalta, E. N.},
  title        = {Deontic Logic},
  booktitle    = {The Stanford Encyclopedia of Philosophy},
  edition      = {Fall 2021},
  year         = {2021}
}

@inproceedings{sordoni2025,
  author       = {Sordoni, A. and Proulx, J. and Gupta, M. and Maillard, J. and Pineau, J.},
  title        = {Are language models deontologically aligned?},
  booktitle    = {ACL 2025},
  year         = {2025}
}

\appendix
\onecolumn
\section{Appendix A: Base-X Addition Repository}

This appendix provides the complete NormCode repository for the base-X addition case study described in Section 8.

\subsection{Input Specification}

The input to the addition plan is a simple JSON structure specifying number pairs with their tensor axes:

\begin{verbatim}
{
  "{number pair}": {
    "data": [["%(123)", "%(98)"]],
    "axes": ["number pair", "number"]
  }
}
\end{verbatim}

The \texttt{\%()} syntax wraps raw values as perceptual signs, deferring their interpretation until the MVP step when an inference actually needs to operate on the data.

\subsection{NormCode Plan}

\subsubsection{Natural Language Translation (.ncn Format)}

For verification by domain experts, the plan translates to a readable narrative. The notation uses (OUTPUT) to indicate what each step produces, (ACTION) to describe how production occurs, (INPUT) to identify data flowing in, (YIELDS) to specify extracted results, (TIMING) to indicate dependencies on prior steps, (CONDITION) to specify when actions execute, and bracketed numbers like [1.2.3] to provide flow indices serving as unique addresses for debugging.

\begin{verbatim}
[1] (OUTPUT) The new number pair
    (ACTION) is obtained by iterating through every number pair in the collection.
    (MECHANISM) Each iteration carries forward the current carry-over number.
    
    [1.1] (OUTPUT) The result of each iteration
        (ACTION) is assigned from the remainder.
        
        [1.1.2] (OUTPUT) The digit sum
            (ACTION) is computed by summing the unit place values of all numbers 
                     together with the current carry-over number.
            (INPUT 1) All unit place values of the numbers in the current pair.
            (INPUT 2) The current carry-over number (initially 0).
            (YIELDS) The sum.
            
            [1.1.2.4] (OUTPUT) All unit place values of numbers
                (ACTION) is collected by grouping across the current number pair.
                
                [1.1.2.4.2] (OUTPUT) Each unit place value
                    (ACTION) is obtained by iterating through each number 
                             in the current pair.
                    
                    [1.1.2.4.2.1] (OUTPUT) The loop result
                        (ACTION) is assigned from the single unit place value.
                        
                        [1.1.2.4.2.1.2] (OUTPUT) The single unit place value
                            (ACTION) is extracted by getting the unit place 
                                     digit of the number.
                            (INPUT) The current number from the pair.
                            (YIELDS) The extracted digit.

        [1.1.3] (OUTPUT) The number pair collection
            (ACTION) is updated by appending a new number pair.
            (CONDITION) This append happens ONLY IF NOT all numbers are zero,
                        AND IF the carry-over number is zero.
            
            [1.1.3.2] (OUTPUT) The number pair to append
                (ACTION) is constructed by iterating through each number 
                         in the current pair.
                
                [1.1.3.2.1] (OUTPUT) The loop result
                    (ACTION) is assigned from the number with last digit removed.
                    
                    [1.1.3.2.1.2] (OUTPUT) The number with last digit removed
                        (ACTION) is computed by: if the number is less than 10, 
                                 output 0; otherwise, remove the unit place digit.
                        (INPUT) The current number.
                        (YIELDS) The shifted number.

            [1.1.3.3] (OUTPUT) The proposition "all numbers are 0"
                (ACTION) is evaluated by checking if each number in the pair 
                         to append is 0.
                (TIMING) This check occurs after the number pair to append is ready.
                (CONDITION) True if ALL numbers satisfy: the number is 0.

            [1.1.3.4] (OUTPUT) The proposition "carry-over number is 0"
                (ACTION) is evaluated by checking if the carry-over number is 0.
                (TIMING) This check occurs after the number pair to append is ready.
                
                [1.1.3.4.2] (OUTPUT) The updated carry-over number
                    (ACTION) is computed from the previous carry-over and new quotient.
                    
                    [1.1.3.4.2.2] (OUTPUT) The new carry-over value
                        (ACTION) is found by computing the quotient of the digit 
                                 sum divided by 10.
                        (TIMING) This occurs after the digit sum is available.
                        (INPUT) The digit sum.
                        (YIELDS) The quotient (new carry).

        [1.1.4] (OUTPUT) The remainder
            (ACTION) is computed by getting the remainder of the digit sum 
                     divided by 10.
            (TIMING) This occurs after the digit sum is available.
            (INPUT) The digit sum.
            (YIELDS) The remainder (the digit to output).

    (INPUT) The initial number pair collection.
\end{verbatim}

\subsubsection{Formal Syntax (.ncd Format)}

The formal NormCode syntax uses structured markers for precise machine interpretation:

\begin{verbatim}
{new number pair} | 1. quantifying
    <= *every({number pair})%:[{number pair}]@(1)^[{carry-over number}<*1>] 
       | 1.1. assigning
        <= $.({remainder}) 
        
        <- {digit sum} | 1.1.2. imperative
            <= ::(sum {1}<$([all {unit place value} of numbers])%_> 
                  and {2}<$({carry-over number}*1)%_> 
                  to get {3}?<$({sum})%_>)
            <- {sum}?<:{3}>
            <- {carry-over number}*1<:{2}> 
            <- [all {unit place value} of numbers]<:{1}> | 1.1.2.4. grouping
                <= &across({unit place value}:{number pair}*1)
                <- {unit place value} | 1.1.2.4.2. quantifying
                    <= *every({number pair}*1)%:[{number}]@(2) 
                       | 1.1.2.4.2.1. assigning
                        <= $.({single unit place value})
                        <- {single unit place value} | 1.1.2.4.2.1.2. imperative
                            <= ::(get {2}?<$({unit place value})%_> 
                                  of {1}<$({number})%_>)
                            <- {unit place digit}?<:{2}>
                            <- {number pair}*1*2
                    <- {number pair}*1

        <- {number pair}<$={1}> | 1.1.3. assigning
            <= $+({number pair to append}:{number pair})%:[{number pair}] 
               | 1.1.3.1. timing
                <= @if!(<all number is 0>) | 1.1.3.1.1. timing
                    <= @if(<carry-over number is 0>)

            <- {number pair to append}<$={1}> | 1.1.3.2. quantifying
                <= *every({number pair}*1)%:[{number}]@(3) 
                   | 1.1.3.2.1. assigning
                    <= $.({number with last digit removed}) 
                    <- {number with last digit removed} 
                       | 1.1.3.2.1.2. imperative
                        <= ::(output 0 if {1}<$({number})%_> is less than 10, 
                              otherwise remove {2}?<$({unit place digit})%_> 
                              from {1}<$({number})%_>) 
                        <- {unit place digit}?<:{2}> 
                        <- {number pair}*1*3<:{1}>
                <- {number pair}*1

            <- <all number is 0> | 1.1.3.3. judgement
                <= :%(True):<{1}<$({number})%_> is 0> | 1.1.3.3.1. timing
                    <= @after({number pair to append}<$={1}>)
                <- {number pair to append}<$={1}><:{1}>

            <- <carry-over number is 0> | 1.1.3.4. judgement
                <= :%(True):<{1}<$({carry-over number})%_> is 0> 
                   | 1.1.3.4.1. timing
                    <= @after({number pair to append}<$={1}>)
                <- {carry-over number}*1 | 1.1.3.4.2. grouping
                    <= &across({carry-over number}*1:
                               {carry-over number}*1<--<!_>>)
                    <- {carry-over number}*1 | 1.1.3.4.2.2. imperative
                        <= ::(find the {1}?<$({quotient})%_> 
                              of {2}<$({digit sum})%_> divided by 10) 
                           | 1.1.3.4.2.2.1. timing
                            <= @after({digit sum})
                        <- {quotient}?<:{1}>
                        <- {digit sum}<:{2}>

        <- {remainder} | 1.1.4. imperative
            <= ::(get the {1}?<$({remainder})%_> 
                  of {2}<$({digit sum})%_> divided by 10) 
               | 1.1.4.1. timing
                <= @after({digit sum})
            <- {remainder}?<:{1}>
            <- {digit sum}<:{2}>

    <- {number pair}<$={1}>
\end{verbatim}

\subsection{Concept Repository}

The concept repository defines over 50 concepts. Ground concepts with pre-populated references include \texttt{\{number pair\}} of type \texttt{\{\}} with reference data \texttt{[["\%(123)", "\%(98)"]]}, representing the input number pairs; \texttt{\{carry-over number\}*1} of type \texttt{\{\}} with reference data \texttt{["\%(0)"]}, representing the initial carry-over state; and \texttt{\{unit place digit\}?} of type \texttt{\{\}} with reference data describing ``1 digit counting from the right,'' representing a semantic specification for extraction.

Functional concepts define the operations. The outer loop \texttt{*every(\{number pair\})\%:[\{number pair\}]@(1)\^{}[\{carry-over number\}<*1>]} is of type \texttt{*every} and iterates with carry state. The grouping operation \texttt{\&across(\{unit place value\}:\{number pair\}*1)} is of type \texttt{\&across} and collects digits across numbers. The imperative \texttt{::(sum \{1\}<...> and \{2\}<...> to get \{3\}?<...>)} is of type \texttt{::(\{\})} and performs digit summation. The judgement \texttt{:\%(True):<\{1\}<\$(\{number\})\%\_> is 0>} is of type \texttt{<\{\}>} and checks if a value is zero.

\subsection{Inference Repository}

The inference repository contains 23 inference entries with their working interpretations.

The outer loop inference at flow index 1 has sequence type ``quantifying,'' infers concept \texttt{\{new number pair\}}, uses the function concept \texttt{*every(\{number pair\})...}, takes value concepts \texttt{[\{number pair\}]}, and context concepts \texttt{[\{number pair\}*1, \{carry-over number\}*1]}. Its working interpretation specifies the ``every'' marker with quantifier index 1, loop base concept, current loop base concept with the *1 marker, group base name, in-loop concept mapping, and concepts to infer.

The digit summation inference at flow index 1.1.2 has sequence type ``imperative\_python,'' infers concept \texttt{\{digit sum\}}, and takes three value concepts. Its working interpretation specifies value ordering (positions 1, 2, 3 for the inputs), enables thinking mode, and indicates non-relation output.

The conditional termination inference at flow index 1.1.3.1.1 has sequence type ``timing,'' infers concept \texttt{@if!(<all number is 0>)}, and uses the function concept \texttt{@if(<carry-over number is 0>)}. Its working interpretation specifies the ``if'' marker and the condition to check.

\subsection{Repository Files}

The complete repository files are available in the codebase at \texttt{infra/examples/add\_examples/repo/}. The \texttt{addition\_concepts.json} file contains 935 lines defining all concepts. The \texttt{addition\_inferences.json} file contains 656 lines defining all inferences with their working interpretations. The \texttt{addition\_inputs.json} file contains 12 lines specifying input data. The Python runner \texttt{ex\_add\_complete\_12\_base.py} provides validation and execution harness.

\end{document}